\documentclass[10pt,twocolumn,letterpaper]{article}

\usepackage{cvpr}              % To produce the CAMERA-READY version
\usepackage{graphicx}
\usepackage{amsmath,bm}
\usepackage{mathtools}
\usepackage{amssymb}
\usepackage{booktabs,tabularx}
\usepackage[inline]{enumitem}
\usepackage{comment}
\usepackage{placeins}
\usepackage{multirow}
\usepackage{flushend}

\usepackage[pagebackref,breaklinks,colorlinks]{hyperref}

\usepackage[capitalize]{cleveref}
\crefname{section}{Sec.}{Secs.}
\Crefname{section}{Section}{Sections}
\Crefname{table}{Table}{Tables}
\crefname{table}{Tab.}{Tabs.}

\def\cvprPaperID{10103} % *** Enter the CVPR Paper ID here
\def\confName{CVPR}
\def\confYear{2023}

\begin{document}

\title{Make-A-Story: Visual Memory Conditioned Consistent Story Generation}

\author{Tanzila Rahman$^{1,3}$ \qquad Hsin-Ying Lee$^{2}$ \qquad Jian Ren$^{2}$ \qquad Sergey Tulyakov$^{2}$  \\ 
Shweta Mahajan$^{1,3}$ \qquad Leonid Sigal$^{1,3,4}$\\
$^1$University of British Columbia \qquad $^2$Snap Inc.  \\
$^3$Vector Institute for AI \qquad
$^4$Canada CIFAR AI Chair \\
%\texttt{trahman8@cs.ubc.ca, my.yang@mail.utoronto.ca,  lsigal@cs.ubc.ca}
}

\iffalse
\author{First Author\\
Institution1\\
Institution1 address\\
{\tt\small firstauthor@i1.org}
\and
Second Author\\
Institution2\\
First line of institution2 address\\
{\tt\small secondauthor@i2.org}
}
\fi
% \maketitle

%%%%%%%%% ABSTRACT
\makeatletter
\DeclareRobustCommand\onedot{\futurelet\@let@token\@onedot}
\def\@onedot{\ifx\@let@token.\else.\null\fi\xspace}

\def\eg{\emph{e.g}\onedot} \def\Eg{\emph{E.g}\onedot}
\def\ie{\emph{i.e}\onedot} \def\Ie{\emph{I.e}\onedot}
\def\cf{\emph{cf}\onedot} \def\Cf{\emph{C.f}\onedot}
\def\etc{\emph{etc}\onedot} \def\vs{\emph{vs}\onedot}
\def\wrt{w.r.t\onedot} \def\dof{d.o.f\onedot}
\def\etal{\emph{et al}\onedot}
\makeatother

\twocolumn[{
\renewcommand\twocolumn[1][]{#1}
\vspace*{-10mm}
\maketitle

\begin{center}
    \centering
    \vspace{-0.2in}
    \includegraphics[width=0.93\textwidth]{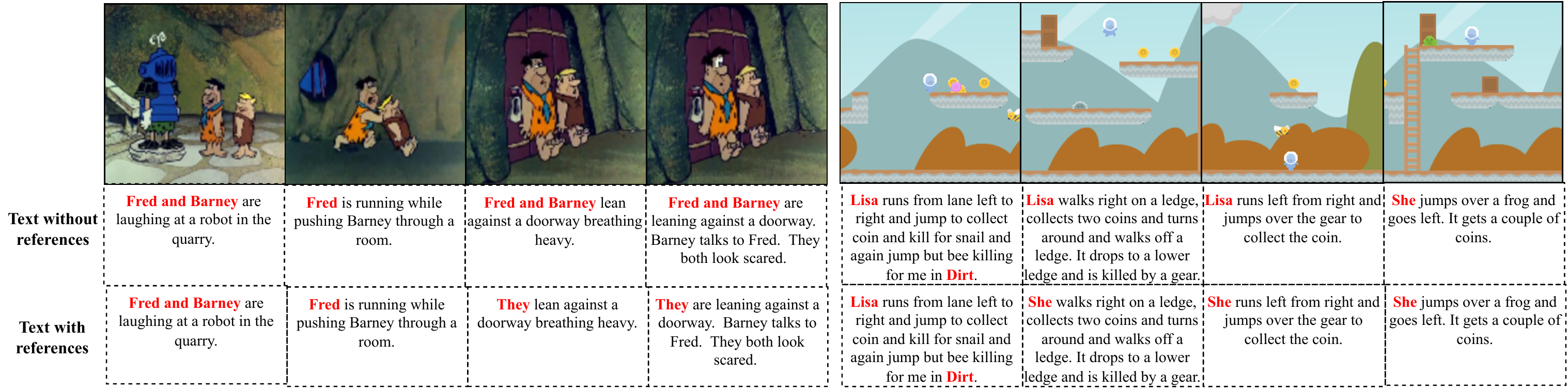}
    % \vspace*{5mm}
    \vspace{-0.1in}
    \captionof{figure}{\textbf{Referential and consistent story visualization.} Examples of more {\em natural} stories with references for the FlintstonesSV~\cite{gupta2018imagine} and MUGEN~\cite{hayes2022mugen} datasets (bottom text) compared to more typical but less natural story text (top). We extend the MUGEN dataset by introducing additional two characters (\emph{e.g.} Lisa and Jhon) and four backgrounds (\emph{e.g.} Sand, Grass, Stone and Dirt). %The visualization of the stories with references, produced by the proposed visual memory conditioned diffusion method, are on top, above each corresponding sentence. 
    % Besides, for natural story progression we introduce co-references of actors and backgrounds in given textual content.
    }
    \label{fig:intro-1}
\end{center}
}]

% \begin{figure*}
%  \centering
%  \includegraphics[scale=0.56]{Figure/intro-1.pdf}
%  \caption{\textbf{Examples of referencing sentences for FlintstonesSV~\cite{gupta2018imagine} and MUGEN~\cite{hayes2022mugen} Dataset.} We extend MUGEN dataset by introducing additional two characters (\emph{e.g.} Lisa and Jhon) and four backgrounds (\emph{e.g.} Sand, Grass, Stone and Dirt). Besides, for natural story progression we introduce co-references of actors and backgrounds in given textual content.}
%  \label{fig:intro-1}
%\end{figure*}

%\begin{figure*}
%  \centering
%  \includegraphics[scale=0.56]{Figure/intro-1.pdf}
%  \caption{\textbf{Examples of referencing sentences for FlintstonesSV~\cite{gupta2018imagine} and MUGEN~\cite{hayes2022mugen} Dataset.} We extend MUGEN dataset by introducing additional two characters (\emph{e.g.} Lisa and Jhon) and four backgrounds (\emph{e.g.} Sand, Grass, Stone and Dirt). Besides, for natural story progression we introduce co-references of actors and backgrounds in given textual content.}
%  \label{fig:intro-1}
% \end{figure*}

\begin{abstract}
There has been a recent explosion of impressive generative models that can produce high quality images (or videos) conditioned on text descriptions. However, all such approaches rely on conditional sentences that contain unambiguous descriptions of scenes and main actors in them. Therefore employing such models for more complex task of story visualization, where naturally references and co-references exist, and one requires to reason about when to maintain consistency of actors and backgrounds across frames/scenes, and when not to, based on story progression, remains a challenge.  In this work, we address the aforementioned challenges and propose a novel autoregressive diffusion-based framework with a visual memory module that implicitly captures the actor and background context across the generated frames. 
Sentence-conditioned soft attention over the memories enables effective reference resolution and learns to maintain scene and actor consistency when needed. To validate the effectiveness of our approach, we extend the MUGEN dataset~\cite{hayes2022mugen} and introduce additional characters, backgrounds and referencing in multi-sentence storylines. Our experiments for story generation on the MUGEN, the PororoSV~\cite{li2019storygan} and the FlintstonesSV \cite{gupta2018imagine} dataset show that our method not only outperforms prior state-of-the-art in generating frames with high visual quality, which are consistent with the story, but also models appropriate correspondences between the characters and the background. % across the frames. 

% Deep learning approaches have been considerably successful in generating high quality images from textual captions. 
% Yet, the extension of these approaches to video or story generation remains challenging.
%  %Compared to text-to-image synthesis, these tasks encompass long narratives and sequential images.
% The goal of this task in addition to generating high quality visual frames in agreement with the sentences, is also to ensure temporal consistency between the frames.
%  Prior work on story generation can synthesize frames with high visual fidelity but is limited in resolving character references and background information across the frames.
%  In this work, we address the aforementioned challenges and propose a novel framework with a visual memory module to capture the character and background consistency across the frames given the text. 
%  To validate the effectiveness of our approach, we extend the MuGen dataset and introduce multiple characters in the story. 
% Each story consists of multiple sentences with character co-references.
% In addition to this, we also introduce different backgrounds, thereby increasing the complexity of the task for background consistency.
% Our experiments for story generation on the MuGen and the Flintstones dataset show that our method not only outperforms prior state-of-the-art in generating frames with high visual quality which are consistent with the story but also models appropriate correspondences between the characters and the background across the frames of the story.
\end{abstract}
%%%%%%%%% BODY TEXT

\section{Introduction}\label{sec:intro}
\vspace{-1mm}
% \vspace{-0.1in}
%positioning the problem formulation
Multimodal deep learning approaches have pushed the quality and the breadth of conditional generation tasks such as image captioning \cite{MaoXYWY14a,JohnsonKF16,VijayakumarCSSL18,wang2018every,mahajan2020diverse} and text-to-image synthesis \cite{ReedAYLSL16,ZhangGMO19,abs-1904-01480,zhu2019dm,zhang2021cross,kang2020contragan}. 
Owing to the technical leaps made in generative models, such as generative adversarial networks (GANs) \cite{GoodfellowPMXWOCB14}, variational autoencoders (VAEs) \cite{KingmaW13} and the more recent diffusion models \cite{ho2020denoising}, approaches for text-to-image synthesis can now generate images with high visual fidelity representative of the textual descriptions. 
The captions, however, in such cases,  are generally short self-contained sentences representing the high-level semantics of a scene.
This is rather restrictive in the real-world applications % to stories or videos
\cite{li2018video,li2019storygan,denton2018stochastic} where
fine-grained understanding of object interactions, motion and background information described by multiple sentences becomes necessary.
One such task is that of {\em story generation} or {\em visualization} -- the goal of which is to generate a sequence of illustrative image frames with coherent semantics given a sequence of sentences \cite{li2019storygan,zeng2019pororogan,maharana2021improving,maharana2022storydall}.

Characteristic features of a good visual story is high visual quality over multiple frames; this includes rendering of discernible objects, actors, poses and realistic interactions of those actors with objects and within the scene.
% smooth interactions between the objects or characters between the frames
%and effective modeling of character motion which varies across the frames. 
%based on the text input is one of the crucial and difficult problem in story visualization.
Moreover, for text-based story generation it is crucial to maintain consistency between the generated frames and the multi-sentence descriptions.
% This encompasses the ability to resolve co-references in the text when encoding actor motion or behaviour across frames so as to make it consistent with the sentences in the story. 
%Yet another important aspect in encoding these interactions is the ability to
Not only the actor context, but also the background of the generated story should be in-line with the description demonstrating effortless transition and adaptation to the changing environments within the story \cite{maharana2021integrating}.

%Prior work
Recent advances on the task of story generation have made significant advances along these lines, showing high visual fidelity and character consistency for story sentences that are self-contained and unambiguous (explicitly mentioning characters and the setting each time). 
While impressive, this setup is fundamentally unrealistic. 
Realistic story text is considerably more complex and referential in nature; requiring ability to resolve ambiguity and references (or co-references) through reasoning. 
% entities are explicitly addressed with a name without any co-references. 
% However, this simplification of the problem may not always exist in the real-world scenarios.
As shown in \cref{fig:intro-1}, while the description corresponding to the first frame has an explicit reference to the character names, the (typical) subsequent frame descriptions, provided by human, contain references such as ``\textit{she, he, they}''.
Moreover, while maintaining character consistency, current approaches are limited in preserving, or transitioning through, the background information in agreement with the text (\cf~\cref{fig:visualization-mugen}) \cite{li2019storygan,maharana2022storydall,maharana2021improving}.

In natural language processing (NLP) co-reference resolution in text is an important and core task \cite{mccarthy1995using,aone1995evaluating,kehler1997probabilistic}. While it maybe possible to apply such methods to story text to first resolve ambiguous references and then generate corresponding images using existing story generation approaches, this is sub-optimal. The reason, is that co-reference resolution in the text domain, at best, would only allow to resolve references and maintain consistency across {\em identity} of the character. Appearance across frames would still lack consistency and require some form of visual reasoning. As also noted in \cite{seo2017visual}, reference resolution in the visual domain, or visio-lingual domain, is more powerful.

%Our contributions
In this work, for the first time (to our knowledge), we study co-reference resolution in story generation. 
Prior work \cite{maharana2021improving} offers limited performance when faced with text containing references (see \cref{sec:experiments}).
We address this by 
% We do so by 
proposing a new autoregressive diffusion-based framework with a visual memory module that implicitly captures the actor and background context across the generated frames. 
Sentence-conditioned soft attention over the memories enables effective visio-lingual co-reference resolution and learns to maintain scene and actor consistency when needed. 
Further, given the lack of datasets that contain references and more complex sentence structure, we extend the MUGEN dataset~\cite{hayes2022mugen} and introduce additional characters, backgrounds and referencing in multi-sentence storylines. 

% , a sub-topic well defined in natural language processing. This task is even more challenging and less studied in the visual domain.

\vspace{0.1in}
\noindent
{\bf Contributions.}
Our contributions are three-fold:
\begin{enumerate*}[label=(\roman*), font=\itshape]
\item First, we introduce a novel autoregressive deep generative framework, \emph{Story-LDM}, that adopts and extends latent diffusion models for the task of story generation. As part of Story-LDM, we propose a meticulously designed memory-attention mechanism capable of encoding and leveraging contextual relevance between the part of the story-line that has already been generated, and the current frame being generated based on learned semantic similarity of corresponding sentences. 
% the current sentence description 
% encoding relationships along the temporal dimension by weighting the semantic similarity between the descriptions of the current frame and the part of story-line that has already been parsed. 
Equipped with this, our sequential diffusion model can generate consistent stories by resolving and then capturing temporal character and background context. %  across the generated frames. 
\item Second, to validate our approach for co-reference resolution, and character and background consistency in the visual domain, we extend existing datasets to include more complex scenarios and, importantly, referential text. Specifically, we extend the MUGEN dataset \cite{hayes2022mugen} to include multiple characters and diverse backgrounds. We also modify FlintstonesSV 
\cite{gupta2018imagine} and  PororoSV~\cite{li2019storygan} dataset to include character references. 
These enhancements allow us to increase the complexity of the aforementioned datasets by introducing co-references in the sentences of a story.
% Furthermore, to emphasize the benefits of our approach we apply reference resolution to a more complex FlintstonesSV dataset \cite{gupta2018imagine}. 
\item Finally, to evaluate different approaches for foreground (character) as well as background consistency we propose novel evaluation metrics. Our results on the MUGEN \cite{hayes2022mugen}, the PororoSV~\cite{li2019storygan} and the FlintstonesSV \cite{gupta2018imagine} datasets show that we outperform the prior state-of-the-art on consistency metrics by a large margin.

\end{enumerate*}

% \leon{The contributions need a bit of restructuring I think.}

\section{Related work}

\noindent
{\bf Text-to-image synthesis.} Deep generative models, particularly, generative adversarial networks (GANs)~\cite{GoodfellowPMXWOCB14}, variational autoencoders (VAEs)~\cite{KingmaW13} and normalizing flows~\cite{DinhKB14,DinhSB17,BhattacharyyaMF20} have been applied to multimodal tasks at the intersection of vision and language. Typical such tasks include image captioning \cite{0009SL17,AnejaICCV2019,mahajan2020diverse} and text-to-image synthesis \cite{ReedAYLSL16,ZhangGMO19,abs-1904-01480,MahajanG020,zhu2019dm,tseng2020retrievegan}.
Early work on text conditioned image synthesis built upon the success of GANs~\cite{ReedAYLSL16}.
% in generating realistic images \cite{ReedAYLSL16}. 
More recent approaches have utilized multi-stage generators  \cite{ZhangGMO19} and normalizing flow-based priors \cite{MahajanG020} in the latent space to model the distribution of images given text. 
%Yin \etal \cite{abs-1904-01480} propose a siamese architecture to separately encode high-level and low-level image features and Zhu \etal \cite{zhu2019dm} leverage a dynamic memory module. 
Various approaches have found cross-domain contrastive loss to improve text-to-image generation models \cite{zhang2021cross,kang2020contragan}. 
DALL-E \cite{ramesh2021zero} and Cogview \cite{ding2021cogview} harness the power of transformers \cite{vaswani2017attention} and discrete variational autoencoders (VQ-VAE) \cite{razavi2019generating} yielding very high quality image samples.

More recent are the advances in diffusion models which have revolutionized the domain of image generation \cite{ho2020denoising}. Diffusion models progressively add noise to the data and learn a reverse diffusion process to reconstruct it. 
Nichol \etal \cite{nichol2021glide} adapted diffusion models for text-to-image generation and explore CLIP \cite{RadfordKHRGASAM21} guided generation as well as classifier-free modeling.
Standard diffusion models are employed directly in the high-dimensional pixel space and therefore, cannot directly be used for the more complex task of story generation.
Recent work \cite{rombach2022high,gu2022vector,cheng2022sdfusion} instead use encodings from pre-trained models as input to the diffusion models, thereby reducing the complexity of the task by working in a lower-dimensional space. 
In this work, we build upon this idea and extend it for sequential story generation.

\begin{figure*}[t!]
\begin{subfigure}[b]{0.55\linewidth}
\centering
   \includegraphics[width=\linewidth]{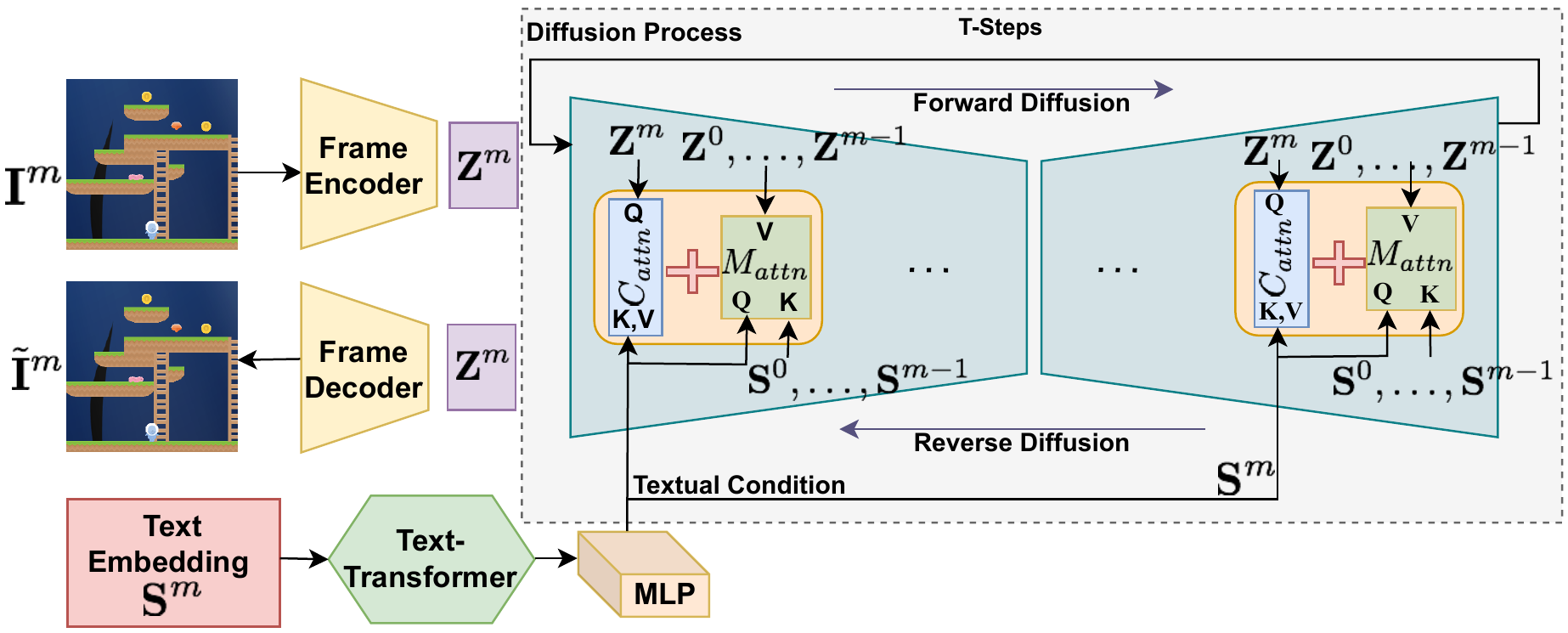}
   \vspace{-2mm}
   \caption{Story-LDM architecture for conditional story generation.}
   \label{fig:architecture}
\end{subfigure}%
\hfill
\begin{subfigure}[b]{0.35\linewidth}
\centering
  \includegraphics[width=\linewidth]{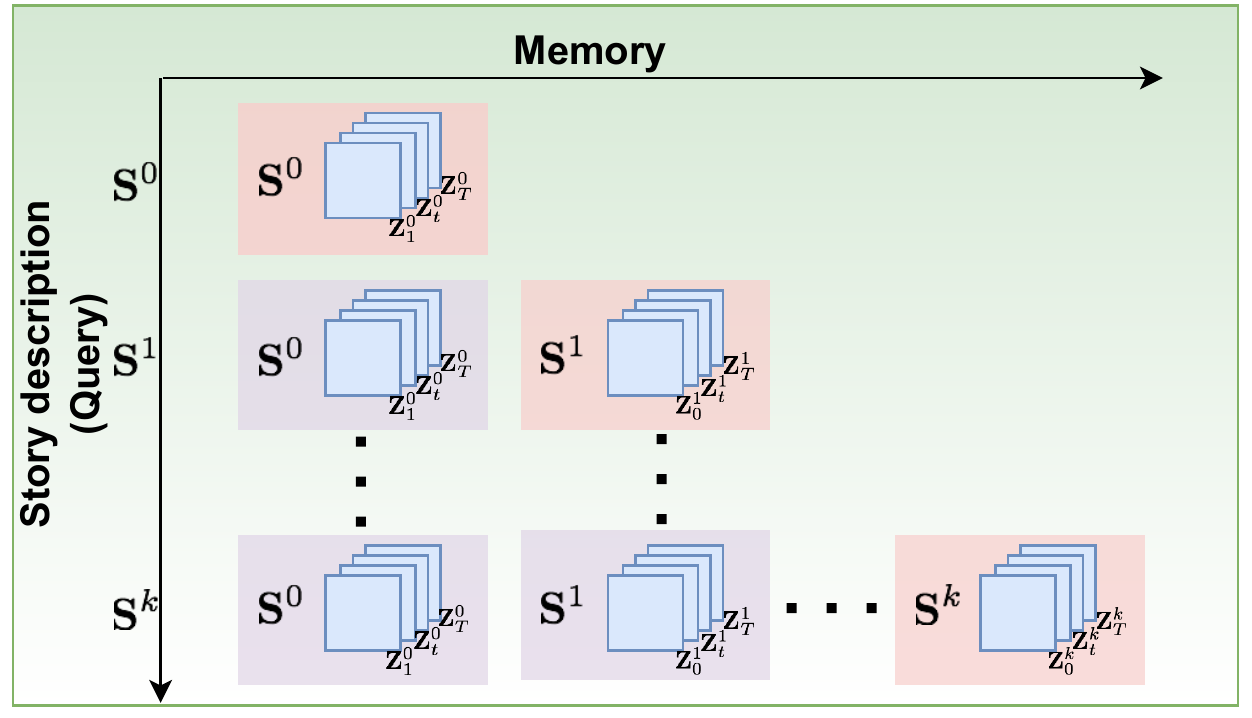}
  \caption{Memory state with query and values for our memory-attention module.}
  \label{fig:key_value}
\end{subfigure}
\vspace{-2mm}
\caption{\textbf{Story-Latent Diffusion Models for consistent story generation.} \emph{(a)} The autoregressive conditional generative Story-LDM model with the proposed memory-attention module. \emph{(b)} A snapshot of the memory for $k$ frames.}\label{fig:archi}
\vspace{-3.5mm}
\end{figure*} 
 
\vspace{0.1in}
\noindent
{\bf Text-to-video synthesis.} One of the challenges of text-to-video synthesis is the smoothness of motion in a video \cite{li2019storygan}. Early work % on text-to-video synthesis 
focused on generating short clips \cite{li2018video,denton2018stochastic}. To effectively learn the motion, various approaches disentangle the motion features from the background information \cite{hao2018controllable,tulyakov2018mocogan,vondrick2016generating}.
Wu \etal \cite{wu2021godiva} propose a novel two-dimensional VQ-VAE and sparse attention module for real-world text-to-video generation.
Singer \etal \cite{singer2022make} decompose the temporal U-Net \cite{RonnebergerFB15} and the attention modules to approximate them in space and time to extend the text-to-image diffusion models to model text-to-video generation.
Ligong \etal \cite{han2022show} propose a transformer framework to jointly model various modalities. 

\vspace{0.0in}
\noindent
{\bf Story Generation.}
Li \etal \cite{li2019storygan} proposed the initial idea and task of story generation. A two-level StoryGAN framework is applied to ensure image-level consistency between each sentence and image pair, and a global discriminator enforces global consistency between the entire image sequence and the story.
Various approaches have proposed improvements to the StoryGAN architecture. Zeng \etal \cite{zeng2019pororogan} introduce sentence-level alignment and word-based attention to improve relevance.
Li \etal \cite{li2020improved} further improve the performance % of the original StoryGAN 
with enhanced discriminators and dilated convolutions.
In \cite{song2020character} foreground-background information is provided as additional supervision and \cite{maharana2021integrating} use video captioning for semantic alignment between text and frames.
Recently, Chen \etal~\cite{chen2022character} adopted visual planning and character token alignment to improve character consistency. 

\vspace{0.07in}
\noindent
{\bf Story Completion.}
Recently, another task for text-to-story synthesis referred to as story completion has been proposed \cite{maharana2022storydall}.
In this task, in addition to sentences, the first frame of the story is provided as input.
In effect, story completion is a simplified variant of story generation. 
% , providing a simplified outlook to the task of story generation. 
StoryDALL-E \cite{maharana2022storydall} leverages models pre-trained for text-to-image synthesis to perform story completion.
Datasets for this task include CLEVR-SV \cite{li2019storygan} and Pororo-SV \cite{gupta2018imagine} which are derived from the CLEVR dataset \cite{johnson2017clevr}, and the Flintstones dataset for text-to-video synthesis has also been modified for the task of story visualization \cite{maharana2021integrating}. 
Additionally, to evaluate the generalization performance, popular DiDeMo dataset for video captioning \cite{anne2017localizing} is adapted for the task in \cite{maharana2022storydall}. 
% \leon{This last sentence does not appear to fit well here.} \\
% \leon{Lets separate story continuation into a separate paragraph.} \\
% \leon{We should maybe have a short paragraph on reference resolution.}

\vspace{0.07in}
\noindent
{\bf Reference Resolution.} Co-reference resolution is an important and well-researched topic in NLP and focuses on resolving the pronouns and their associated entities.
Classic methods in NLP to co-reference resolution employ decision trees \cite{mccarthy1995using,aone1995evaluating}, maximum-entropy modeling \cite{kehler1997probabilistic}, cluster-ranking \cite{rahman2011narrowing} and classification algorithms \cite{soon2001machine}. More recent approaches \cite{fei2019end,joshi2020spanbert,kobayashi2022end} leverage neural network architectures to obtain improved performance with Transformers \cite{devlin2018bert}.
Seo \etal \cite{seo2017visual} proposed visual co-reference resolution for the task of Visual Question-Answering (VQA) dialogs. 
% , an important aspect for performing visual dialog and proposed an attention mechanism that takes into account the best match from the previous attention.
We take inspiration from \cite{seo2017visual}, but propose a much more sophisticated  memory-attention module that allows us to perform visio-lingual co-reference resolution (and visual consistency modeling) for 
visual story generation. 
\section{Approach}
To generate temporally consistent stories based solely on the linguistic story-line, we develop a deep generative approach with autoregressive structure.
We build upon the success of diffusion models in modeling the underlying data distribution of images to produce high quality samples, and learn the generative conditional distribution of the visual story based on the textual descriptions.
Given that the multi-frame stories involve high-dimensional data input, we employ Latent Diffusion Models \cite{rombach2022high}, such that diffusion models can be applied in a computationally efficient manner.
Besides, to ensure temporal consistency and smooth story progression, we propose a novel memory attention mechanism which not only attends to the multimodal representations of the current frame but also takes into account the already generated semantics of the previous frames.
This module also allows us to resolve ambiguous references (\eg, he/she, they, \etc) using visual memory and is the core of our technical contribution. 
We first provide an overview of the Diffusion Models and the Latent Diffusion Models, following which we present our autoregressive latent diffusion model for stories called \emph{Story-LDM}\footnote{\url{https://github.com/ubc-vision/Make-A-Story}}.

\subsection{The Latent Diffusion Model Backbone}

\noindent
{\bf Diffusion Models.} Diffusion models are a class of generative models that approximate the underlying data distribution $p(\mathbf{x})$ by denoising a base (Gaussian) distribution in multiple steps using a reverse process of a fixed Markov Chain of length $T$.
To estimate $p(\mathbf{x})$, the forward diffusion process starts from the input data $\mathbf{x}_0 = \mathbf{x}$ and gradually adds noise to obtain a set of noisy samples ${\mathbf{x}_1, \dots, \mathbf{x}_T}$ such that $\mathbf{x}_T \sim \mathcal{N}(0,1)$ represents a sample from a Gaussian distribution. Under the Markov assumption, the probability of the forward process modeling the distribution $q(\mathbf{x}_{0:T} \mid \mathbf{x}_0)$ and the reverse diffusion process estimating probability at an earlier time-step are formulated as:
\begin{align}
\begin{split}
q(\mathbf{x}_{1:T} \mid \mathbf{x}_0)% &\coloneqq \prod_{t=1}^T q(\mathbf{x}_t \mid \mathbf{x}_{t-1})\\ 
&\coloneqq \prod_{i=1}^{T}\mathcal{N}(\mathbf{x}_t;\sqrt{1-\beta_t}\mathbf{x}_{t-1},\beta_t \bm{I})\\
    p_\theta(\mathbf{x}_{0:T}) &= p_\theta(\mathbf{x}_T) \prod_{i=1}^Tp(\mathbf{x}_{t-1} \mid \mathbf{x}_t).
\end{split}
\end{align} 
Here, $\{\beta_i\}_{i=1}^T$ is the variance schedule for each time-step such that $\mathbf{x}_{T}$ is nearly a Gaussian.
%The reverse diffusion process estimates the probability at an earlier time-step given the current state $\mathbf{x}_t$ under the model parameters $\theta$ as,
%\begin{align}
%    p_\theta(\mathbf{x}_{0:T}) = p_\theta(\mathbf{x}_T) \prod_{i=1}^Tp(\mathbf{x}_{t-1} %\mid \mathbf{x}_t).
%\end{align}
%During training, at any time-step $t$, the model learns the reverse process $p_\theta(x_{t-t} \mid x_t)$ parameterized by $\theta$, that is, given the noisy input 
The model parameters $\theta$ are learnt with the following objective, 
\begin{align}
\mathcal{L}_{DM} \coloneqq \mathbb{E}_{t,\mathbf{x},\epsilon}\left[\|\epsilon - \epsilon_\theta(\mathbf{x}_t,t)\|_2^2\right],
\label{eq:DM:obj}
\end{align}
where $\epsilon \sim \mathcal{N}(0,1)$ and $\epsilon_\theta(\mathbf{x}_t,t)$, $t=1,\ldots, T$ is a sequence of denoising autoencoders with noisy input $\mathbf{x}_t$ predicting the noise that was added to the original input $\mathbf{x}$.

%Popular approaches synthesizing images with diffusion model have yielded state-of-the-art results in terms of image quality and realism making these approaches desirable for visual scene understanding.
Despite yielding state-of-the-art results in various image generation tasks, diffusion models directly operating in the high-dimensional pixel are computationally expensive and resource exhaustive. 
This limits their application to an even higher-dimensional data such as multi-frame stories or video datasets, which is the focus of this work.  

\vspace{0.1in}
\noindent
{\bf Diffusion Models in the Latent Space.} To broaden the applicability of the diffusion models to very high-dimensional data \eg high-resolution images, Latent Diffusion Models (LDM) \cite{rombach2022high} first compress the original image to a lower-dimensional space using perceptual image compression.
An auto-encoder approach is employed such that the original spatial structure of the input image is preserved in the latent space. 
That is, the encoder $E(\cdot)$ maps the input image $\mathbf{I} \in \mathbb{R}^{H \times W \times 3}$ to a latent representation $\mathbf{Z} \in \mathbb{R}^{h \times w \times c}$, downsampling the image to a lower spatial dimension.
Following this, the diffusion model is applied to the latents $\mathbf{Z}$, where time-conditioned U-Net $\epsilon_\theta(\mathbf{Z}_t,t)$ is employed to model the diffusion process.
The objective of the diffusion model from \cref{eq:DM:obj} becomes, 
\begin{align}
\mathcal{L}_{LDM} \coloneqq \mathbb{E}_{t,E(\mathbf{Z}),\epsilon}\left[\|\epsilon - \epsilon_\theta(\mathbf{Z}_t,t)\|_2^2\right],
\label{eq:LDM:obj}
\end{align}
During training, a forward diffusion process is applied to generate $\mathbf{Z}$, which are mapped to the original image space using a decoder $D(\cdot)$.

\subsection{Story-Latent Diffusion Models} 

Given a textural story, characterized by sequence of $M$ sentences $\mathbf{S}_{txt} = \{ \mathbf{S}^0, \ldots, \mathbf{S}^M \}$, the goal of story generation is to produce a sequence of corresponding frames $\mathbf{S}_{img} = \{\mathbf{I}^0, \ldots, \mathbf{I}^M \}$ that visualize the story. 
We note that this is a more difficult problem than one of story continuation~\cite{maharana2022storydall}, where in addition to the textual story $\mathbf{S}_{txt}$ approaches have access to a source frame $\mathbf{I}^0$ for additional context at inference time. 
During training it is assumed that we have access to paired dataset of $N$ samples $\mathcal{D} = \{ \mathbf{S}^{(i)}_{txt}, \mathbf{S}^{(i)}_{img} \}_{i=1}^{N}$. 
% Given a sequence of image frames $I^0, \ldots, I^M$ and the corresponding descriptions $S^0, \ldots, S^M$, we are tasked with the goal of producing coherent and high-quality visual stories based only on the input text.
% Thus we intend to solve a harder generative problem compared to \cite{maharana2022storydall}, where the source frame is provided as an additional input.
We extend latent diffusion models to this task, by allowing them to generate multi-frame stories autoregressively, and by introducing rich conditional structure that takes into account current sentence as well as context from earlier generated frames through visual memory module. This visual memory allows the model to incorporate character/background consistency and resolve text references when needed, resulting in improved performance.

\begin{table}
\scriptsize
\centering
  \begin{tabularx}{\columnwidth}{@{}Xccccc@{}}
  \toprule
    Dataset & \# Ref (avg.) & \# Chars & \# Backgrounds \\
    \midrule 
    MUGEN~\cite{hayes2022mugen} & None & 1 & 2\\
    Extended MUGEN & 3 & 3 & 6 \\
    FlintstonesSV~\cite{gupta2018imagine} & 3.58 & 7 & 323 \\
    Extended FlintstonesSV & 4.61 & 7 & 323\\
    PororoSV~\cite{li2019storygan} & 1.01 & 9 & None\\
    Extended PororoSV & 1.16 & 9 & None\\
    \bottomrule
  \end{tabularx}
  \vspace{-2mm}
  \caption{\textbf{Dataset statistics of the MUGEN, FlintstonesSV and PororoSV}. }
  \vspace{-5mm}
  \label{table:st_dataset}
\end{table}

\begin{table*}[t!]
%\scriptsize
  \begin{tabularx}{\textwidth}{@{}cXcccccc@{}}
  \hline
    Dataset&Method & w/ ref. text & Char-acc ($\uparrow$)   & Char-F1 ($\uparrow$)   & BG-acc ($\uparrow$)   & BG-F1 ($\uparrow$)   & FID  ($\downarrow$)  \\
    \toprule 
    \multirow{4}{*}{\rotatebox[origin=c]{90}{{Flintstones}}}&VLCStoryGAN~\cite{maharana2021integrating} & $\times$ & 27.73 & 42.01 & 4.83 & 16.49 & 120.85 \\
    &LDM~\cite{rombach2022high} & $\times$  & 79.86 & 92.33 & 48.02 & 37.86 & 61.40 \\
    &LDM~\cite{rombach2022high} & \checkmark & 57.38 & 78.68 & 44.19 & 28.25 & 87.39 \\
    &Story-LDM (Ours) & \checkmark &  69.19 & 86.59 & 35.21 & 28.80 & 69.49 \\
    %& \bfseries Story-LDM (Ours) (new) & \checkmark &  80.91 & 93.37 & 51.31 & 43.75 & 66.09 \\

    \iffalse
    \midrule
    \parbox{2.1mm}{\multirow{4}{*}{\rotatebox[origin=c]{90}{\scriptsize{PororoSV}}}}&DUCO-STORYGAN~\cite{maharana2021improving} & \checkmark & 13.97 & 38.01 & - & - & 96.51 \\
    &VLCStoryGAN~\cite{maharana2021integrating} & \checkmark & 17.36 & 43.02 & - & - & 84.96 \\
    &LDM~\cite{rombach2022high} & \checkmark  & 16.59 & 56.30 & - & - & 60.23 \\
    &Story-LDM (Ours) & \checkmark & \bfseries 20.26 & \bfseries 57.95 & - & - &\bfseries 36.64 \\
    %& \bfseries Story-LDM (Ours) (new) & \checkmark &  19.66 &  53.06 & - & - & 31.53 \\
    \fi
    
    \midrule
    \parbox{2.1mm}{\multirow{2}{*}{\rotatebox[origin=c]{90}{\scriptsize{MUGEN}}}}&LDM~\cite{rombach2022high} & \checkmark  & 31.39 & 21.28 & 15.74 & 18.66 & 120.99 \\
    &Story-LDM (Ours) & \checkmark & \bfseries 93.40 & \bfseries 95.60 & \bfseries 92.19 & \bfseries 92.37 &\bfseries 62.16 \\
    \bottomrule
  \end{tabularx}
  \vspace{-2mm}
  \caption{{\bf Quantitative results.} Experimental results on the FlintstoneSV and the MUGEN datasets.}
  \label{table:ex-flintstone}
  \vspace{-5mm}
\end{table*}

Given a condition $\mathbf{y}$, LDM utilizes a cross-attention layer with key $(\mathbf{K})$, query $(\mathbf{Q})$ and value $(\mathbf{V})$ where,
\begin{align}
    \text{Attention}(\mathbf{K},\mathbf{Q},\mathbf{V}) = \text{softmax}\left(\frac{\mathbf{Q}\mathbf{K}^T}{\sqrt{d}}\right).\mathbf{V}.
    \label{eq:LDM:attention}
\end{align}
Here, $\mathbf{Q} = \mathbf{W}_{Q}.\hat{f}(\mathbf{Z})$, $\mathbf{K}=\mathbf{W}_K.f(\mathbf{y})$ and $\mathbf{V}=\mathbf{W}_V.f(\mathbf{y})$, and $\mathbf{W}_Q \in \mathbb{R}^{d \times d_q}$, $\mathbf{W}_K \in \mathbb{R}^{d \times d_k}$ and $\mathbf{W}_V \in \mathbb{R}^{d \times d_v}$ are learnable parameters, $\hat{f}(\mathbf{Z})$ an intermediate flattened feature representation of $\mathbf{Z}$ within the diffusion model and $f(\mathbf{y})$ the feature representation of the condition $\mathbf{y}$.
The objective in \cref{eq:LDM:obj} for conditional generation becomes,
\begin{align}
\mathcal{L}_{LDM} \coloneqq \mathbb{E}_{t,E(\mathbf{I}),\epsilon}\left[\|\epsilon - \epsilon_\theta(\mathbf{Z}_t,f(\mathbf{y}),t)\|_2^2\right].
\label{eq:LDM:cond_obj}
\end{align}
Note that the denoising autoencoders $\epsilon_\theta$ now additionally depend on the condition encoding $f(\mathbf{y})$.

For a sample $\{\mathbf{S}^{(i)}_{txt}, \mathbf{S}^{(i)}_{img}\}$, we first project all the $M$ input frames of the story, $\mathbf{I}^0, \ldots, \mathbf{I}^M$ onto a low-dimensional space using a frame-encoder $E$, and obtain
the encoded frames $\mathbf{Z}^0, \ldots, \mathbf{Z}^M$ for a single story\footnote{We drop the superscript denoting sample $i$ for ease of notation.}.
To effectively condition on the corresponding frame, we apply this cross-attention layer to the intermediate representations of the neural network for each frame $\mathbf{I}^m$ and its textual description  $\mathbf{S}^m$ using  \cref{eq:LDM:attention}, where the relevance of the description $\mathbf{S}^m$ is weighted by the similarity between the textual representation and the encoded frame representation $\mathbf{Z}^m$ (\cf~\cref{fig:architecture}).

For sequential generation, the model in addition to the current state, requires information from all the previous states.
To enable this, the diffusion process for any frame representation $\mathbf{Z}^m$ is conditioned on the visual representations of the previous frames $\mathbf{Z}^0,\ldots,\mathbf{Z}^{m-1}$ as well as the sentence descriptions $\mathbf{S}^0,\ldots,\mathbf{S}^{m}$. 
This conditioning is realized through a novel \emph{Memory-attention} module which forms the basis of our autoregressive approach.

\begin{figure*}[th!]
  \centering
  \includegraphics[scale=0.58]{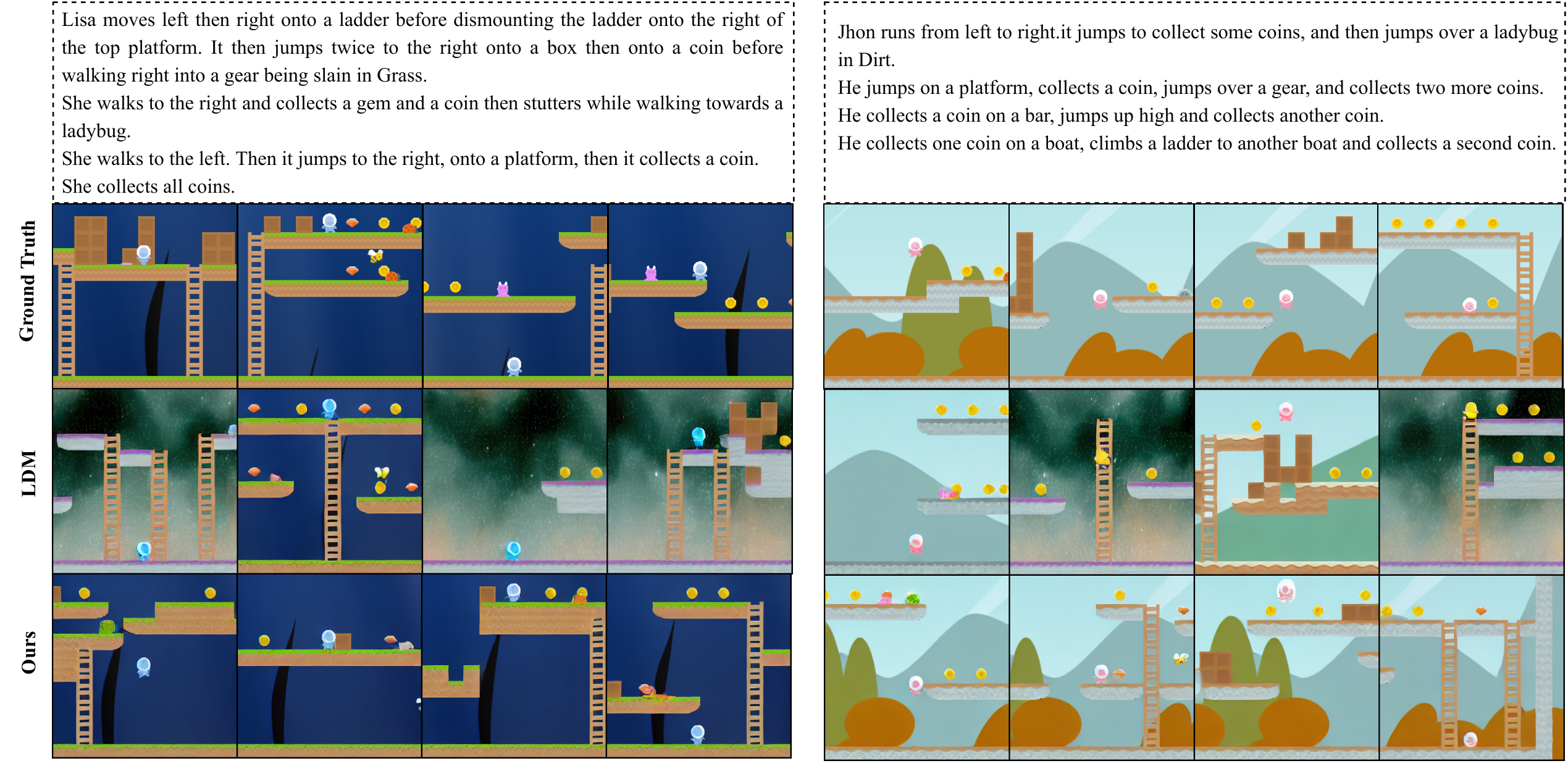}
  \vspace{-2mm}
  \caption{\textbf{Story generation result for MUGEN.} Here we can compare between our method and LDM~\cite{rombach2022high}. See text for details.}
  \vspace{-2mm}
  \label{fig:visualization-mugen} 
\end{figure*}
\begin{figure*}[h!]
  \centering
  \includegraphics[scale=0.58]{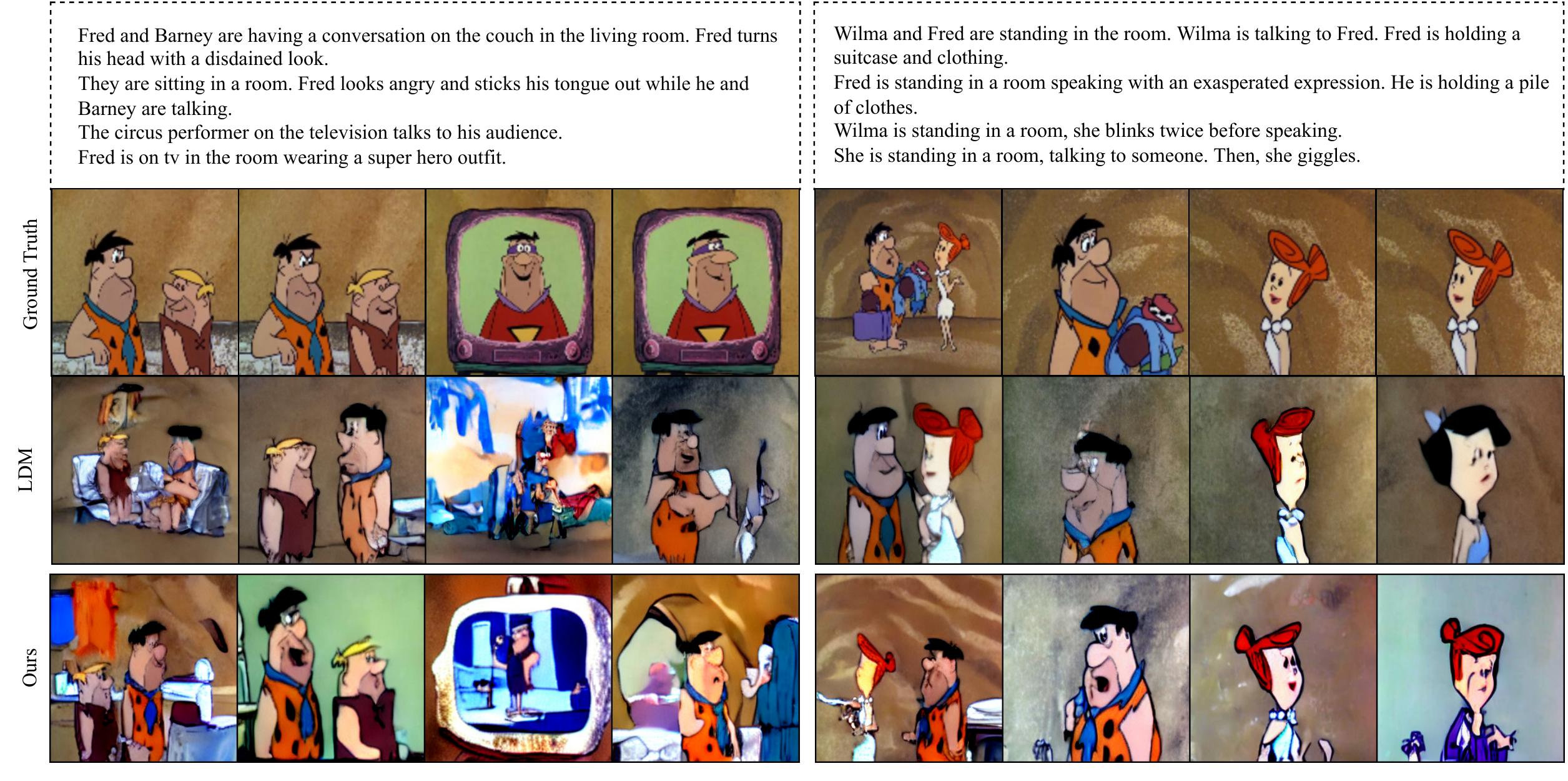}
  \vspace{-2mm}
  \caption{\textbf{Story generation results for FlintstoneSV.} Our method is able to generate more consistent characters/backgrounds.}
  \vspace{-5mm}
  \label{fig:visualization-flintstone} 
\end{figure*}

\paragraph{Memory-attention Module.}
To capture the spatio-temporal interactions across multiple frames and sentences for a story, in our conditional diffusion model, we condition the frame $\mathbf{Z}^m$ not only on the corresponding text $\mathbf{S}^m$ but also on the previous texts $\mathbf{S}^i$, for $i \in \{0, \ldots m-1\}$.
This conditioning is applied throughout the $T$ time-steps of the diffusion process for $\mathbf{Z}^m$.
The conditional denoising autoencoder thus models the conditional distribution $p(\mathbf{Z}^m \mid \mathbf{Z}^{<m}, \mathbf{S}^{\leq m})$.
The (conditional) generative process of our \emph{Story-LDM} approach over the $T$ steps of the diffusion process for a single frame is thus given by
\begin{align}
\begin{split}
    p(\mathbf{Z}^m|\mathbf{Z}^{<m}, \mathbf{S}^{\leq m}) = p&(\mathbf{Z}^m_T|\mathbf{Z}^{<m}, \mathbf{S}^{\leq m} )\\
    &\prod_{i=1}^{T}p(\mathbf{Z}^m_{i-1}| \mathbf{Z}^m_{i}, \mathbf{Z}^{<m},  \mathbf{S}^{\leq m}).
    \label{eq:storyldm:singleframe}
\end{split}
\end{align}
%With this factorization, we condition the
The key motivation for this approach is to propagate the semantic (visual and textual) features from the already processed story-line based on the relevance of the current description to the previous frames as well as the previous descriptions.
We achieve this by implementing a special attention layer, called \emph{memory-attention module}. Similar to the cross-attention layer, we utilize the attention mechanism based on the key, query and value formulation. In this case \cref{eq:LDM:attention} becomes,
\begin{align}
\begin{split}
    \mathbf{Q} &= \mathbf{W}_{Q}.f(\mathbf{S}^m), \mathbf{K}=\mathbf{W}_K.f(\mathbf{S}^{<m}),\\ \mathbf{V} &=\mathbf{W}_V. \hat{f}({\mathbf{Z}}^{< m}),
\end{split}
\end{align}
where $\hat{f}(.)$ is applied to align the dimensions of the values, $\mathbf{V}$ with the keys, $\mathbf{K}$.
In the memory attention module, the relevance the query $\mathbf{Q}$ which depends on the current sentence $\mathbf{S}^m$ and the keys $\mathbf{K}$ which represent the previous sentences $\mathbf{S}^{< m}$ is used to weight the feature representations $\mathbf{Z}^{<m}$. The aggregated representation now contains the information relevant for the current frame $\mathbf{Z}^m$ from the already generated story-line (see~\cref{fig:key_value}).
That is, our mechanism based on the similarity of the current sentence to the previous sentences in the story, identifies the features in the previous frames which are of importance to the context of the current frame. This may include recurrence of certain semantics with-in the story such as characters or backgrounds.
In all, this formulation of the diffusion process allows us to maintain temporal consistency as we amplify the visual feature information from the sequence of story already generated.
%The model is thus capable of cross-referencing the semantic information both in the visual domain and the textual domain. 
This allows the model to implicitly capture temporal dependencies in storylines for resolving ambiguities in character and background information.

Given the above conditioning, the objective for the story latent diffusion model  for a single frame is formalized as
\begin{align}
    \mathcal{L}_{story-LDM} =  \mathbb{E}_{\mathbf{Z}, \epsilon,t}\left[\|\epsilon - \epsilon_{\theta^m}(\mathbf{Z}^m_t,\mathbf{S}^{\leq m},\mathbf{Z}^{<m},t)\|_2^2\right],
\end{align}
where $\epsilon_{\theta^m}$ are the denoising autoencoders for the frame $m$.

Having formalized the diffusion process for single frame generation, the generative process for the entire story-line using \cref{eq:storyldm:singleframe} for autoregressive conditional frame generation, is given by
\begin{align}
    p(\mathbf{Z}^{0:M} \mid \mathbf{S}^{0:M}) = p(\mathbf{Z}^0 \mid \mathbf{S}^0) \prod_{i=m}^{M}p(\mathbf{Z}^m \mid \mathbf{Z}^{<m}, \mathbf{S}^{\leq i}).
    \vspace{-0.2cm}
\end{align}

Notably, the conditioning is applied to the all states within the diffusion process \ie, for all $\mathbf{Z}^m_t$, $t \in \{1, \ldots T\}$ at each diffusion step, we apply the cross attention as well as the memory attention module allowing us effectively capture the temporal context. 
%Model name could be consistent latent diffusion model (Consistent-LDM) or Story aware latent diffusion model (Story-LDM) or autoregressive latent diffusion model (autoregressive-LDM). 

\vspace{-0.05in}
\paragraph{Network Architecture.} To generate visual storylines, in our \emph{Story-LDM}, we first introduce an autoregressive structure and modify the two-dimensional U-Net in \cite{rombach2022high} to so as to process the temporal information\  in the storyline. 
As shown in \cref{fig:architecture}, the frame encoder, $E$ equipped with the positional information of the frame in the sequence, is applied to get the low-dimensional representation $\mathbf{Z}^m$ for all the frames in a datapoint from $\mathcal{D}$.
A text-based transformer is applied to get a suitable representation for the sentence $\mathbf{S}^m$. The U-Net is then applied to model the diffusion process over $T$ time-steps. 
The layers within the U-Net are augmented with the cross-attention layer and our memory attention layer. After each downsampling or upsampling operation we apply the attention mechanism to reinforce the conditioning on the already encoded (learned) story-line up to the previous time-step.
For any frame $m$, the cross-attention $C_{attn}$ is given by
\begin{align}
    C_{attn} = \sum_i \hat{f}(\mathbf{Z}^m)_if(\mathbf{S}^m)_i
\end{align}
where $\hat{f}(\mathbf{Z}^m)$ and $f(\mathbf{S}^m)$ are the representations of the frame encoding $\mathbf{Z}^m$ within the neural network and sentence $\mathbf{S}^m$ respectively such that they have same dimensions.
Similarly, the memory attention $M_{attn}$ is computed as,
\begin{align}
    M_{attn} =& \sum_{k=1}^{m-1}\sum_i \hat{f}(\mathbf{Z}^k)_if(\mathbf{S}^k)_if(\mathbf{S}^m)_i
\end{align}
The output of the attention-module is then computed as the aggregation, $C_{attn}+M_{attn}$.

Starting from the noise sample $\mathbf{Z}^m_T$ the output of the reverse diffusion process $\mathbf{Z}^m_0$ is reconstructed using the frame decoder $D$ to get the final image. 
Having outlined the details of our Story-LDM framework, we show through extensive experiments on the task on story generation,  the effectiveness and the benefits of our powerful conditioning based on memory-attention.

\section{Datasets and Evaluation Metrics}\label{sec:datasets}

In this paper, we formulate story generation with  co-references to actors and backgrounds across frames. 

\vspace{0.07in}
\noindent
\textbf{Datasets.}
Since reference resolution has not been studied in story generation, to validate our approach on this much harder task, we construct the following datasets:
\begin{enumerate*}[label=(\roman*), font=\itshape]
\item We take an existing story-generation dataset --   FlintstonesSV~\cite{gupta2018imagine}, and modify the sentences by replacing the named entities (characters) with references where possible; including pronouns such as {\em he}, {\em she}, or {\em they} (\cf~\cref{fig:intro-1}).
This dataset contains $ 20132$-training, $2071$-validation and $2309$-test stories with $7$ main characters and $323$ backgrounds. 

\item MUGEN~\cite{hayes2022mugen} is a video dataset collected from the open-sourced platform game CoinRun~\cite{cobbe2019quantifying}.
The dataset is divided into  $ 104,796$-train and $11,802$ test stories with 96 to 602 frames. %of 3.2s to 21s duration (96 to 602 frames). 
We extend the MUGEN dataset by introducing two additional characters \emph{Lisa} and \emph{Jhon} (we rename \emph{Mugen} to \emph{Tony}). 
We construct stories of four frames and corresponding text, ensuring consistent co-referencing in the story; each story has 3 such references. % In the story-line with four frames and descriptions, the extended version of the dataset has on-average 3 references per story.
Moreover, we augment the existing two backgrounds (\emph{Planet} and \emph{Snow}) with four additional backgrounds: \emph{Sand}, \emph{Dirt}, \emph{Grass} and \emph{Stone}.

\item We also modify existing PororoSV~\cite{li2019storygan} dataset which contains 10191/2334/2208 train/val/test set. Similarly, we reference characters by pro-nouns to generate more natural story.

\end{enumerate*}
We show in \cref{fig:intro-1}, example stories from the two modified datasets and enlist the complete statistics in \cref{table:st_dataset}.

\vspace{0.02 in}
\noindent
\textbf{Evaluation Metrics.} To measure the consistency of the characters as well as the backgrounds in the generated stories, we consider following evaluation metrics:
\begin{enumerate*}[label=(\roman*), font=\itshape]
  \item \emph{Character Classification}: Following~\cite{maharana2021integrating}, we consider fine-tuned Inception-v3 to measure the classification accuracy and F1-score. Frame accuracy evaluates the character match to the ground-truth and F1-score measures the quality of generated characters in the predicted images.
  
  \item \emph{Background Classification}: Similar to character classification, we use fine-tuned Inception-v3 to measure the correspondence of the background to the ground-truth and consider F1-score as a measure of quality. 
  
  \item \emph{Frechet Inception Distance (FID)}:  To assess the quality of images,we consider FID score~\cite{heusel2017gans} which is the distance between feature vectors from real and generated images.
\end{enumerate*}
%The goal of our proposed method is to generate a consistent story where current text is co-referencing actors/backgrounds from previous text. Therefore, the task is different from story generation~\cite{maharana2021integrating} or story continuation~\cite{maharana2022storydall}. Hence, we consider LDM~\cite{rombach2022high}\footnote{\url{https://github.com/CompVis/latent-diffusion}} to generate text-to-image based story without using our proposed auto-regressive memory modules as our baselines for both MUGEN and FlintstonesSV datasets.

\begin{figure}[t]
  \centering
  \includegraphics[scale=0.50]{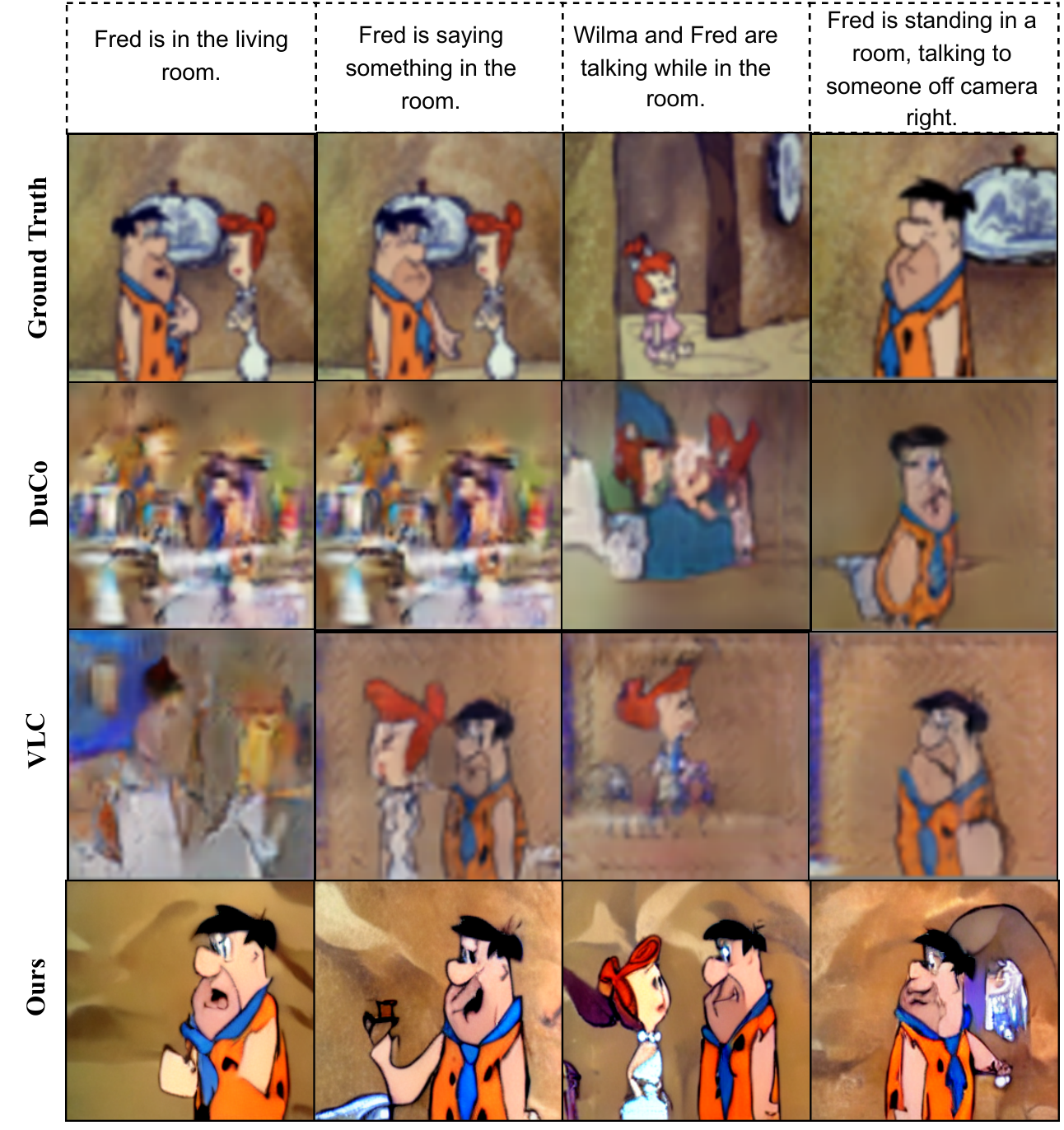}
    \vspace{-0.1in}
  \caption{{\bf Qualitative Comparison on Story Generation.} Comparison on the FlintstonesSV dataset visual story generation.}
  \label{fig:visualization-flintstone-vlcstroygan}
    \vspace{-0.15in}
\end{figure}
\begin{figure}[th!]
  \centering
  \includegraphics[width=0.8\columnwidth]{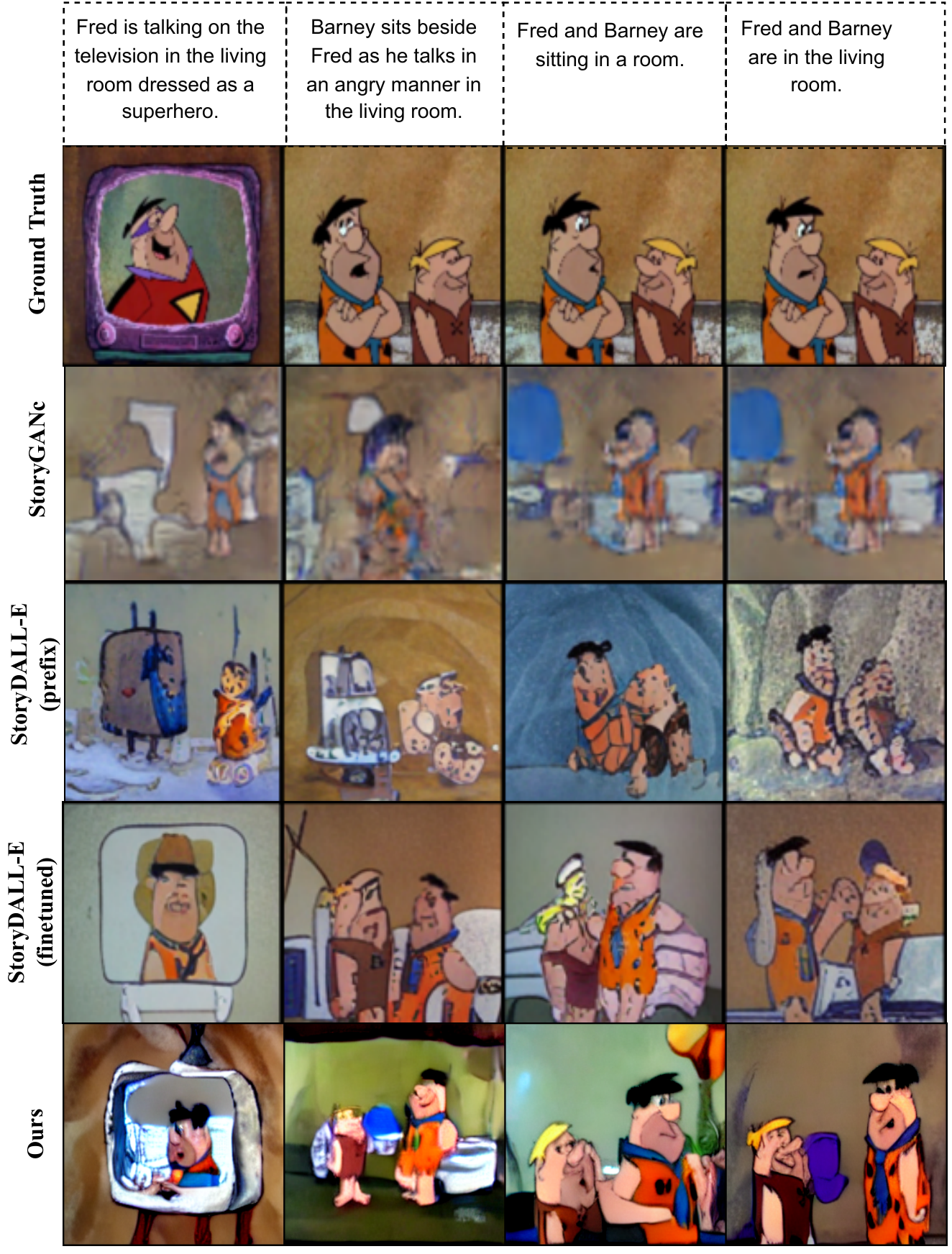}
    \vspace{-0.1in}
  \caption{{\bf Comparison to Story Continuation.} Comparison on the FlintstonesSV dataset for story continuation. Prior work uses first frame as additional input to the model; our model does not.}
  \label{fig:visualization-flintstone-storyDalle}
  \vspace{-0.2in}
\end{figure}

\section{Experiments}\label{sec:experiments}

\iffalse
\begin{table}
\scriptsize
\centering
  \begin{tabularx}{\columnwidth}{@{}Xccccc@{}}
  \midrule
    Method & Char-acc  & Char-F1  & BG-acc  & BG-F1  & FID  \\
    \toprule 
    %StoryGANc(BERT) & 55.78 & 70.45 & - & - & 91.37 \\
    %StoryGANc (CLIP) & 58.39 & 72.80 & - & - & 90.29 \\
    %StoryDALL-E(prompt) & 32.54 & 42.48 & - & - & 53.71  \\
    %StoryDALL-E(fine-tune) & 55.19 & 73.43 & - & - & 26.49  \\
    VLCStoryGAN~\cite{maharana2021integrating} & 22.93 & 37.78 & 2.27 & 9.42 & 237.28 \\
    LDM~\cite{rombach2022high} & 35.65 & 60.82 & \textbf{29.39} & 18.49 & 87.39 \\
    \midrule 
    Ours & \textbf{43.19} & \textbf{64.62} & 25.95 & \textbf{22.82} & \textbf{74.04} \\
    \bottomrule
  \end{tabularx}
  \caption{Experimental results on FlintstoneSV Dataset.}
  \label{table:ex-flinstone}
  \vspace{-0.1in}
\end{table}

\begin{table}
\scriptsize
\centering
  \begin{tabularx}{\columnwidth}{@{}Xccccc@{}}
  \toprule
    Method & Char-acc  & Char-F1 & BG-acc & BG-F1 & FID \\
    \midrule 
    LDM~\cite{rombach2022high} (Original Text) & \textbf{45.65} & \textbf{68.99} & \textbf{34.79} & \textbf{26.98} & \textbf{61.40} \\
    LDM~\cite{rombach2022high} (Ref. Text) & 35.65 & 60.82 & 29.39 & 18.49 & 87.39 \\
    \bottomrule
  \end{tabularx}
  \caption{Comparison between original text and our modified referencing text on FlintstoneSV dataset.}
  \label{table:ex-comparison-flinstone}
\end{table}
\fi

In this section, we evaluate our Story-LDM approach for consistent story generation with reference resolution.

\vspace{0.02in}
\noindent
\textbf{Baselines.}  We construct a strong baseline with the LDM\footnote{\url{https://github.com/CompVis/latent-diffusion}}~\cite{rombach2022high} which contains a cross-attention layer to generate text-to-image based story, without using our proposed autoregressive memory modules as our baselines for MUGEN, PororoSV and FlintstonesSV datasets. 
The parameters of the diffusion model within the Story-LDM are initialized with the pre-trained LDM~\cite{rombach2022high}. 
Similarly, for the textual embedding, we use BERT-tokenizer~\cite{devlin2018bert} and use the pre-trained text-transformer from LDM. %with the pre-trained weights from LDM. %We do not fine-tune the frame autoencoder from LDM on MUGEN/FlintstonesSV datasets.

\vspace{0.02in}
\noindent
\textbf{Quantitative Results.} Table~\ref{table:ex-flintstone} shows quantitative results for consistent story generation on the FlintstoneSV dataset. We compare the performance of our approach (row 4) to the LDM~\cite{rombach2022high} which we train/test with both original (row 2) as well as the co-referenced (row 3) descriptions.  Furthermore, we include the results of the state-of-art VLCStoryGAN~\cite{maharana2021integrating} (row 1) with the original text of the dataset\footnote{Results for \cite{maharana2021integrating} were obtained using pretrained model provided by original authors in private communication.} (\ie~without co-references).
We note that VLCStoryGAN was shown to be better than Duco-StoryGAN \cite{maharana2021improving}, CP-CSV \cite{song2020character} and original StoryGAN \cite{li2019storygan} (see \cite{maharana2021integrating}).

Based on Table~\ref{table:ex-flintstone} we make three observations:
(1) Our LDM baseline is better than VLCStoryGAN on the original reference-free text (\cf~\cref{table:ex-flintstone}, rows 1 \& 2).
(2) Reference resolution makes the task considerably harder. With the reference text in our modified dataset, we observe a drop in performance in terms of character and background classification scores (\cf~\cref{table:ex-flintstone}, rows 2 \& 3).  
(3) Our model, with memory-attention module, significantly outperforms the baseline (\cf~\cref{table:ex-flintstone}, rows 3 \& 4) both in terms of generative image quality and character consistency; and outperforms SoTA of VLCStoryGAN by $\sim 41\%$ percentage points on character accurary (while performing a more difficult version of the task). Further, our model, that is required to conduct reference resolution, comes close to the LDM trained with original, reference-free, text (\cf~\cref{table:ex-flintstone}, rows 3 \& 4), which can be viewed as a sort of an upper bound.
% on possible performance.

% To show that we are solving a much harder task with reference resolution, we compare the results of the LDM baseline with and without the referenced descriptions. With the referenced text of our modified dataset, we observe a drop in performance in terms of character and background classification scores (\cf~\cref{table:ex-flintstone}, rows 2 \& 3).
% Our LDM baseline also outperforms the VLCStoryGAN~\cite{maharana2021integrating} without the reference text.
% With this stronger LDM approach, we compare our approach on the co-referenced text where our method offers competitive performance. 

%On the PororoSV dataset, our method outperforms previous baseline models (including LDM baseline) in character evaluation metrics. To be noted, PororoSV~\cite{li2019storygan} dataset has no background information, therefore we only perform character level evaluation on this dataset.

On the MUGEN dataset, our method outperforms the strong LDM baseline with gains of $\sim 62\%$ on character accuracy and  $\sim 76\%$ on the background accuracy, thereby showing the advantages of the memory-attention mechanism for consistent story generation. We note that MUGEN dataset has more references across story scenes. Flintstones while contains more references per story overall, many of those references are within scenes as opposed to across scenes. Meaning that in terms of reference impact on consistency, MUGEN dataset is actually harder. Experimental results on the PororoSV dataset are provided in the Supplemental.

\vspace{0.02in}
\noindent
\textbf{Qualitative Results.} \Cref{fig:visualization-mugen} illustrates qualitative results on the MUGEN dataset. Rows 1, 2 \& 3 show ground truth, LDM~\cite{rombach2022high} and our Story-LDM approach,  respectively. 
Here, we see that our method is able to maintain consistency in terms of both character and background. Similarly, in \Cref{fig:visualization-flintstone} we can show the results on FlintstoneSV dataset which further validates the strong performance of our method when generating high-quality, consistent story. 
Compared to the LDM, our approach is able to adapt to the diverse backgrounds in the story descriptions.

\begin{figure}
  \centering
  \includegraphics[width=0.78\columnwidth]{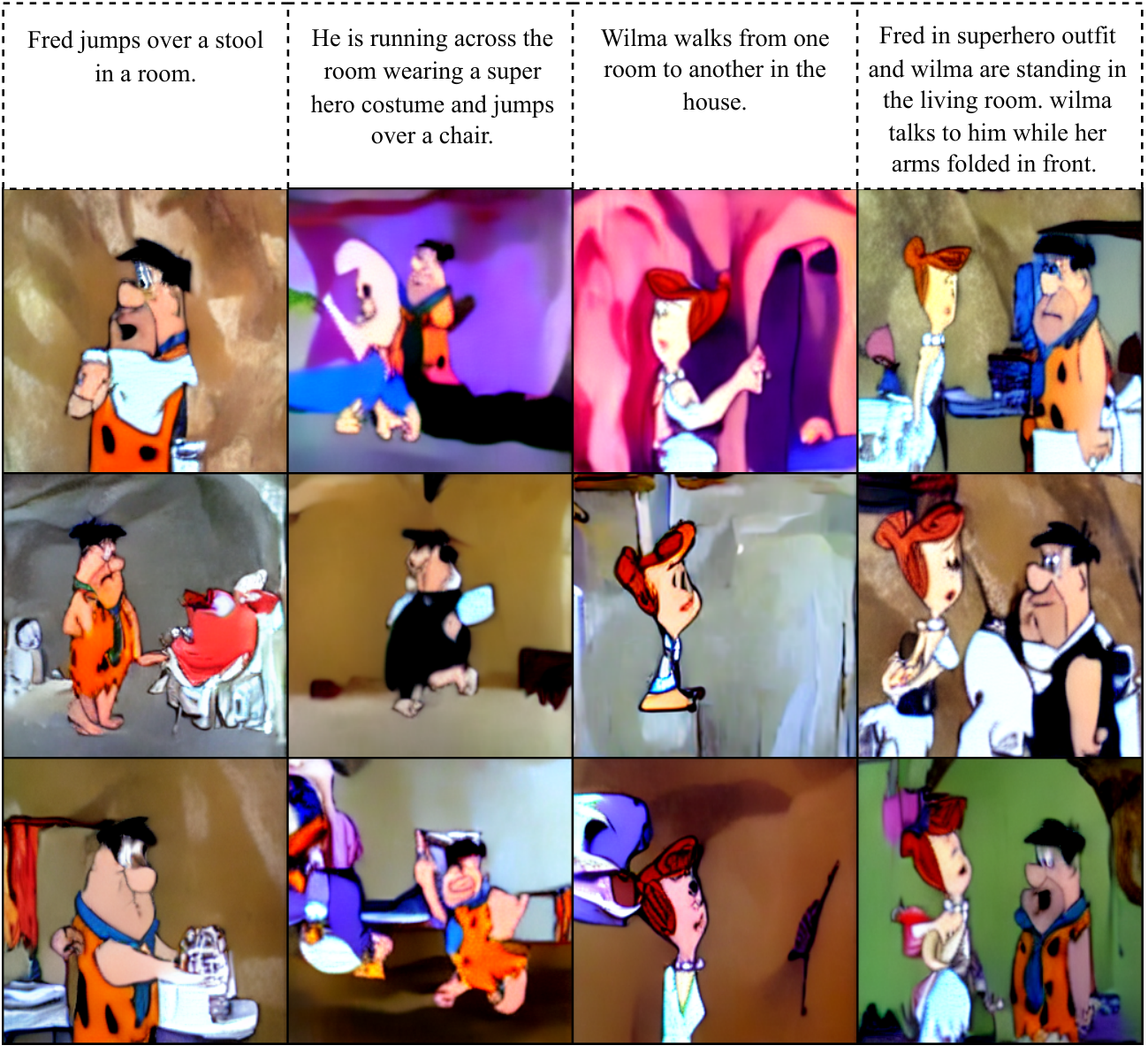}
    \vspace{-0.1in}
  \caption{{\bf Story Diversity.} Diverse outputs for a single storyline obtained with our Story-LDM.}
  \label{fig:diverse_output}
  \vspace{-0.20in}
\end{figure}

\begin{figure}
  \centering
  \includegraphics[width=0.78\columnwidth]{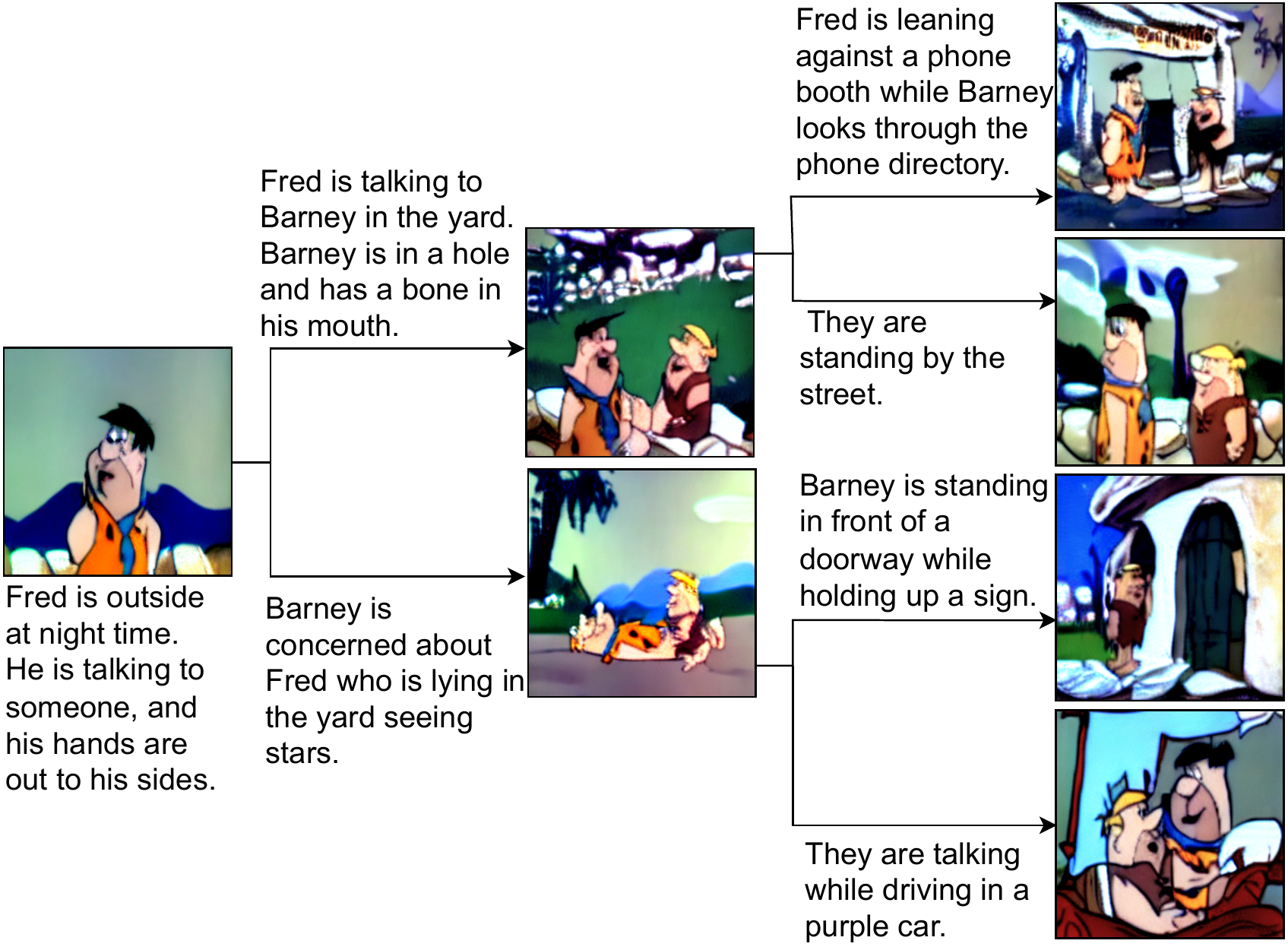}
  \vspace{-0.15in}
  \caption{{\bf Branching Storyline.} Generating different yet consistent stories by branching the storyline. Frames in the later columns are generated based on earlier ones and corresponding text.} 
  % second column are generated by conditioning on the first frame/text and the frames in the third column are generated conditioned on the first and second frames and their texts.}
  \label{fig:branching}
  \vspace{-0.2in}
\end{figure}

\vspace{0.02in}
\noindent
\textbf{Additional Results.} 
We compare the qualitative results of our method to both story generation~\cite{maharana2021integrating} and story continuation~\cite{maharana2022storydall} in \cref{fig:visualization-flintstone-vlcstroygan,fig:visualization-flintstone-storyDalle} respectively. The comparative images are taken directly from respective papers. We note that story continuation \cref{fig:visualization-flintstone-storyDalle} is solving a different (easier) problem and with text that contains no-references.
This makes the comparison to our method, which receives fewer inputs, not very meaningful. % and more anecdotal. 
Nether-the-less, our approach, that can resolve references and is solving a harder story generation task, obtains highly competitive results.
% Here again, we demonstrate the superior performance of our approach compared to the prior state-of-the-art approaches.
%All baseline visualizations are taken from the original paper. 
Furthermore, to show that our autoregressive visual memory module can generate diverse stories conditioned on the current and previous condition, we create different story-lines starting for a single sentence. In \cref{fig:branching}, we can see for reference `they', the model can generate both the characters according to the storyline already parsed. 
Moreover, in \cref{fig:diverse_output} we show that our approach can not only generate consistent visual stories, but also diverse frames for the same text (\cf~\cref{fig:diverse_output}). Additional results are provided in the Supplemental. 


\vspace{-0.05in}

\vspace{-0.05in}
\section{Conclusion}\label{sec:conclusion}

In this paper, we formulate consistent story generation in a more realistic way by co-referencing actors/backgrounds in the story descriptions. 
We develop an autoregressive Story-LDM approach with memory attention capable of maintaining consistency across the frames based on the previously generated frames and their corresponding descriptions.
We introduced modified datasets to evaluate the performance for reference resolution.
We expect our proposed formulation and models to be conductive to the real-world use cases and further the research.
% for consistent and high quality video generation.
%In future, we will extend the work for video based consistent story generation.

%%%%%%%%% REFERENCES
{\small
\bibliographystyle{ieee_fullname}
\bibliography{egbib}
}

\end{document}